\documentstyle[10pt, a4paper, conference, compsocconf, graphicx]{IEEEtran}

\begin{document}

\bibliographystyle{plain}

\title{Genetic Algorithms and the Art of Zen} 
\author{Jack Coldridge and Martyn Amos  \\ \\ Department of Computing and Mathematics \\ Manchester Metropolitan University, \\ Manchester M1 5GD, UK \\ \\ Email: M.Amos@mmu.ac.uk} 
\date{}

\maketitle

\begin{abstract}
In this paper we present a novel genetic algorithm (GA) solution to a simple yet challenging
commercial puzzle game known as the Zen Puzzle Garden (ZPG). We describe the game in detail, before presenting a suitable encoding scheme and fitness function for candidate solutions. We then compare the performance of the genetic algorithm with that of the A* algorithm. Our results show that the GA is competitive with informed search in terms of solution quality, and significantly out-performs it in terms of computational resource requirements. We conclude with a brief discussion of the implications of our findings for game solving and other ``real world" problems.
\end{abstract}

\section{Introduction}

\begin{figure}[]
   \begin{center}
   \includegraphics[width=8cm,height=6cm]{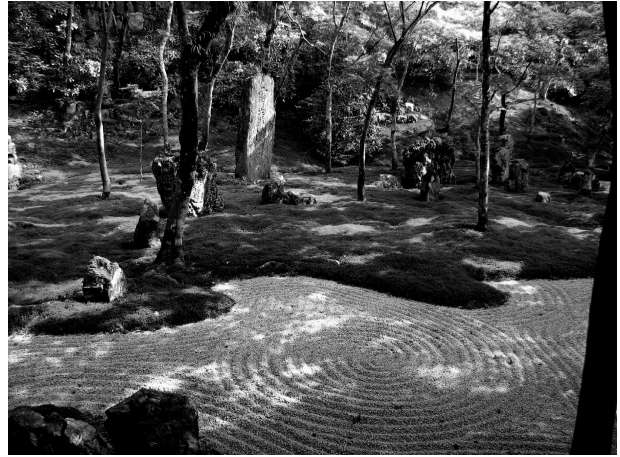}
       \caption{Komyozenji temple garden.}
       \label{fig:kom}
   \end{center}
  \end{figure}

The Zen Puzzle Garden (ZPG) \cite{Lex} is a one-player puzzle game involving
a Buddhist monk raking a sand garden.  It is inspired by Japanese garden design (for example, the Komyozenji temple garden is shown in Figure ~\ref{fig:kom}). One common feature of such gardens is a flat region of sand or small pebbles, which is raked into a pattern. The ZPG is one example of a {\it transport puzzle};  these are problems that involve the player moving entities around a given domain (e.g., boxed around a warehouse), starting at some initial configuration, until they attain pre-defined goal conditions. Entities must move according
to the constraints of the puzzle, and may only move between connected positions (that is, an entity may not be ``lifted" off the board and replaced at a position perhaps far from its initial location). A graphical representation of the problem may use vertices to represent the set of positions an entity may occupy, with connecting edges determined either from any explicitly named connections, or from those implied by arrangement on the board or within a grid.

The rest of the paper is organized as follows: We first describe related work in Section ~\ref{zpg} and give an in-depth description of the problem in Section ~\ref{zpg}, before describing two solution methods (genetic algorithm and A*) for the ZPG in Section ~\ref{algs}. Experimental results are presented in Section ~\ref{results}, before we conclude with a brief discussion of the implications of our findings in terms of broader applicability.

\section{Previous work}

\begin{figure}[h]
   \begin{center}
   \includegraphics[width=8cm,height=5.5cm]{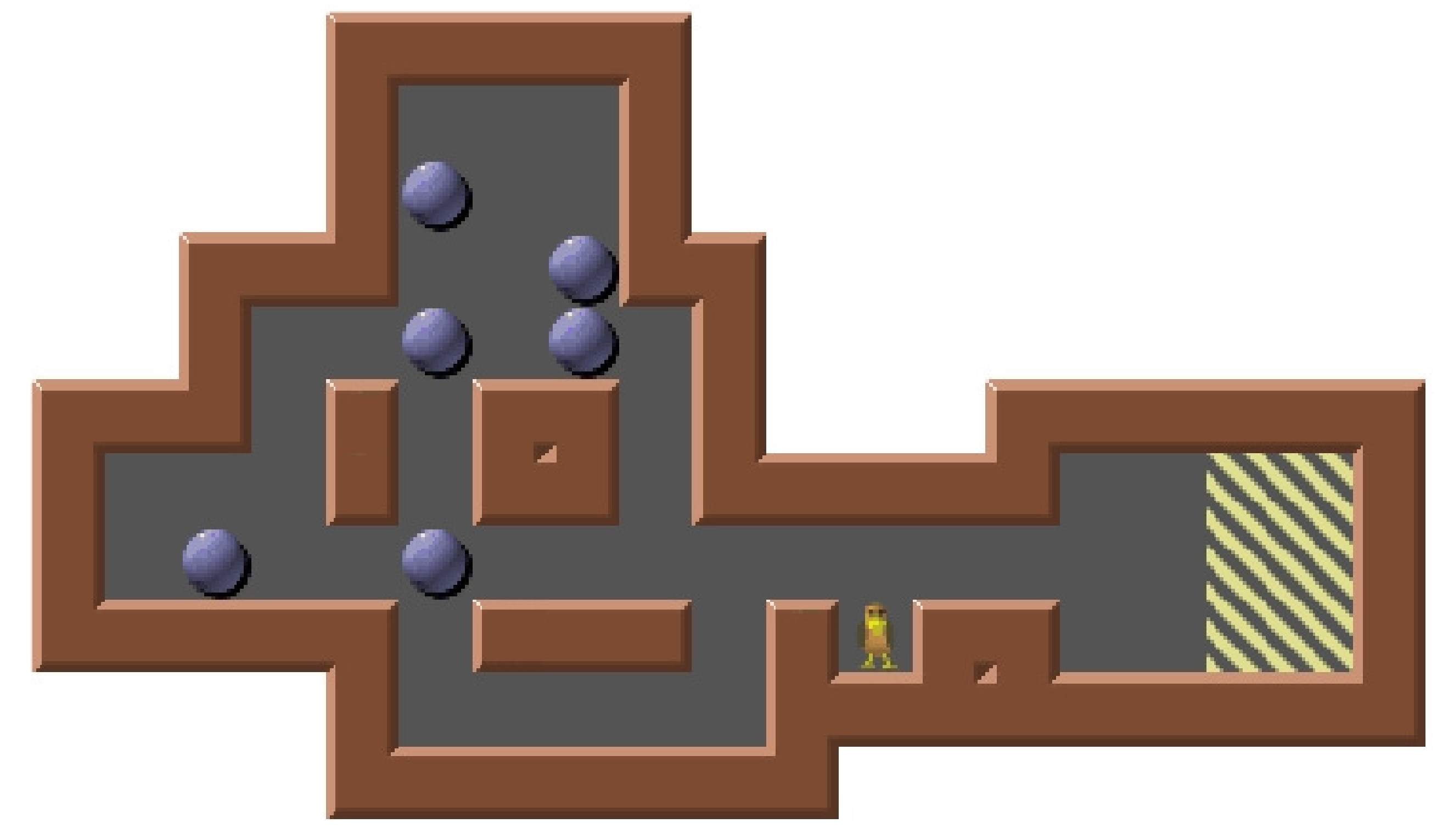}
       \caption{Example Sokoban problem.}
       \label{fig:sokboard}
   \end{center}
 \end{figure}

Many transport puzzles require the player to make a trip around a board between
given start and end positions, and puzzles may be extended via the introduction of objects
or obstacles to the board, which must be collected or moved to satisfy given constraints.
A well-studied example of the transport puzzle is {\it Sokoban} \cite{dor,jung}. In this game the player takes control of a warehouse keeper
whose job it is to push boxes around a maze and into designated target locations;
only one box may be pushed at any one time, and boxes may not be pulled. This challenge of the puzzle is derived from this latter condition, since if a box is pushed into a corner of any construction then the game is lost. An example Sokoban puzzle is depicted in Figure ~\ref{fig:sokboard}, with the ``boxes" depicted as circles, and the target region as a hatched area. Sokoban is known to be {NP}-hard \cite{dor}. Various AI-based techniques have been applied to its solution, including multi-agent systems \cite{berg}, abstraction and decomposition \cite{bot}, embedding domain-specific knowledge \cite{jung2}, and heuristic search \cite{jung3}. The applicability of such methods to ZPG is, however, not clear, since the problem features additional complicating factors (described in the next Section).

According to an in-depth search of the literature, no attempts to automatically solve this particular problem have
been previously documented. We believe the genetic algorithm (GA) \cite{goldberg89} to be a good candidate method for its solution. We therefore describe preliminary work on applying this method to a new variant of the transport problem. We present a comparison between the GA and a ``base-line", search-tree-based method, in the hope that this will motivate further study in the future.

\section{The Zen Puzzle Garden}
\label{zpg}

The ZPG game takes place on a garden board comprised of a two-dimensional grid
of {\it sand} squares surrounded by a {\it perimeter region}. Two of the boards
supplied with the game are partitioned into different sections
separated by path regions; these are not considered here. The
objective of the game is to move a {\it monk} character around the garden, causing him to
completely rake the available surface. The monk always begins a game
at the same point on the perimeter, regardless of the board being
played, and the following rules apply:

\begin{itemize}
\item The monk may move freely around the perimeter region.
\item The monk may only move within the von Neumann neighbourhood (i.e., no diagonal movements are allowed).
\item The perimeter is left by entering the sand on any un-raked (i.e., empty) square.
\item Vacating an empty sand square causes that
square to be permanently raked. The monk cannot move from the
perimeter onto an empty sand square and then immediately back onto
the perimeter, as this is considered a ``step back".
\item Once moving on sand, the monk continues to move in a straight line
until he encounters either the perimeter (in which case he moves
onto it), a raked square or an \emph{object} (in both cases, he
stops moving). The monk may not turn corners while moving. These two
rules are central to the challenge of the game -- if the monk could
be moved on a square-by-square basis then most boards would be
trivial to solve.
\item Moveable objects may only be pushed onto an empty
sand square. They may not be pushed off the garden into the
perimeter region. A single move pushes an ornament one square, if
possible. When pushing an ornament the monk does not continue moving
until he is no longer able to, unlike normal movement.
\item The current game ends if the monk is moved into a position
such that he is in a ``dead end" (i.e., unable to make a legal
move).
\end{itemize}

\begin{figure}[]
   \begin{center}
   \includegraphics[width=8cm]{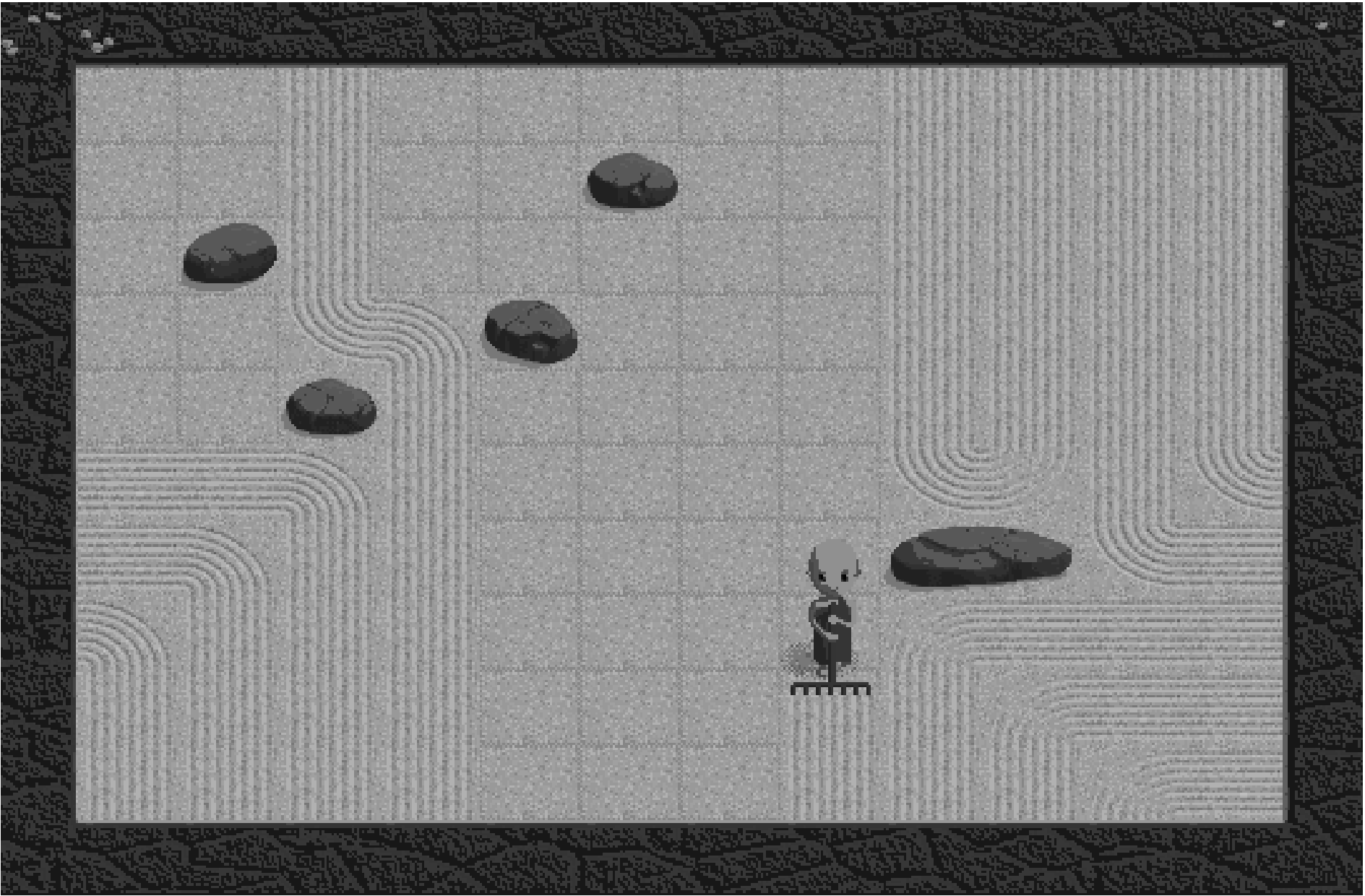}
       \caption{Example Zen Puzzle Garden board.}
       \label{fig:zenboard}
   \end{center}
 ~\\ 
 \end{figure}

\noindent Objects may be one of three types:

\begin{itemize}
\item \emph{Rocks} are placed in a fixed position at the start of the
game, and may not be moved.
\item \emph{Ornaments} also start in the same place at the beginning of each
game. They may only be pushed into an empty square, and not into the
perimeter (see above).
\item \emph{Leaves} are coloured either yellow, orange or red, and must be
collected (i.e., moved over) in that order, at which time they are
removed from the board. An orange or red leaf is classed as an
immoveable object until the preceding leaves have been collected.
\end{itemize}

An example board is depicted in Figure
~\ref{fig:zenboard}. In this situation the monk is half-way through
completing a move, and will completely rake the current column
before moving onto the perimeter.

A puzzle board is completed when all initially empty squares (i.e.,
any sand square not covered by either a rock or an ornament) have
been raked and the monk has stepped onto the perimeter region. This
last condition is necessary, as a square is not considered to be
raked until it has been vacated.

Although a formal proof is beyond the scope of this paper, previous work \cite{dem} suggests that the ZPG is {NP}-complete. The authors study ``pushing block" puzzles (like Sokoban), and introduce variants in which blocks must slide their maximal extent when pushed, and where a player's path must not cross itself. These variants are demonstrated to be {NP}-hard, and future work will use this work to formally establish the intractability of ZPG. 

\section{Two methods for the ZPG}
\label{algs}

In this section we describe two solution methods for the ZPG; a search-tree method and a genetic algorithm (GA). We first give a representation scheme for the game, before describing a generic simulator we built for the game. We then give details of the A* and GA algorithms.

\subsection{Representation}

\begin{figure}[]
   \begin{center}
   \includegraphics[scale=0.4]{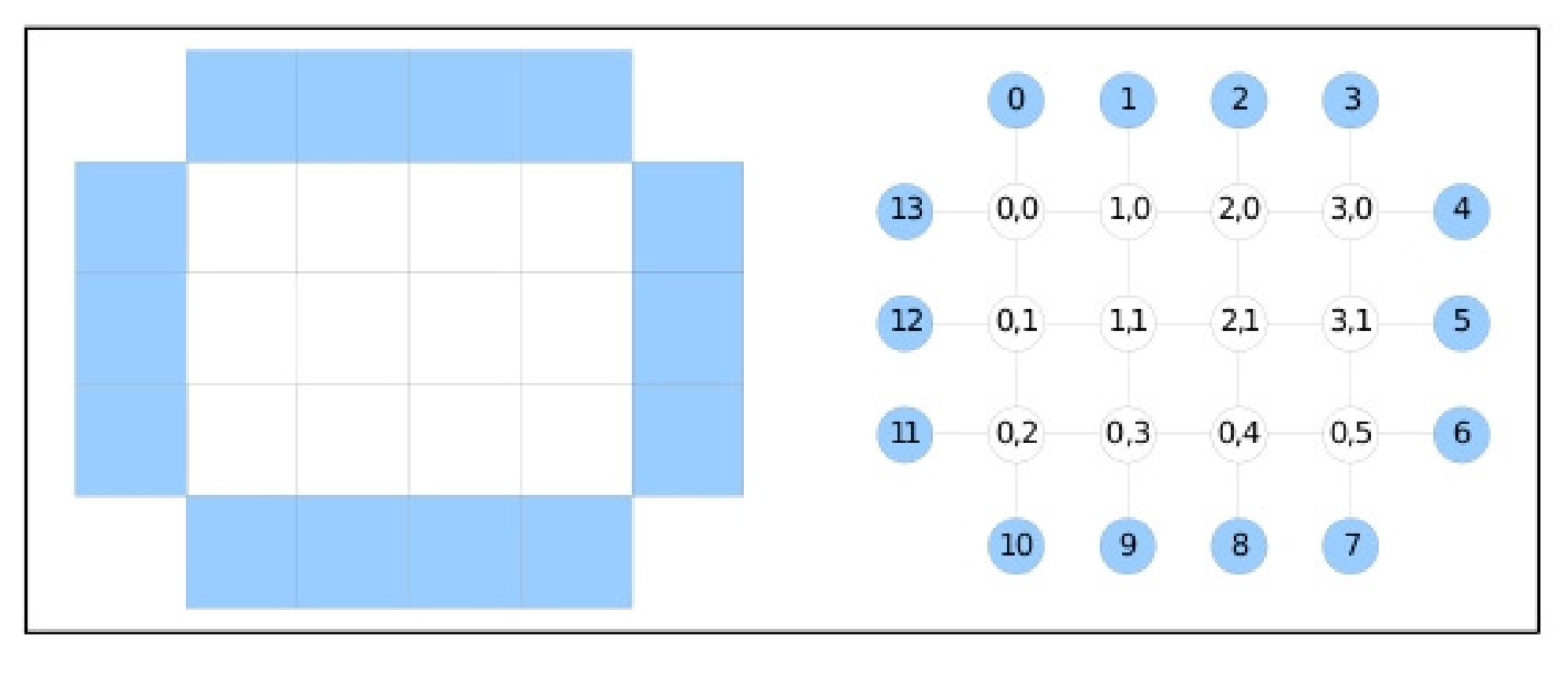}
       \caption{Graph-based representation of ZPG board.}
       \label{fig:graph}
   \end{center}
 ~\\ 
 \end{figure}

In Figure ~\ref{fig:graph}, we show a graph-based representation of the ZPG board. Perimeter tiles are represented by coloured nodes, and sand tiles labelled according to their coordinated on the board. Entities representing the player and obstacles
may occupy vertices on the graph or travel (where allowed) along connecting edges between them. Although not shown in Figure ~\ref{fig:graph}, the monk may move freely between perimeter nodes (that is, the perimeter subgraph is fully connected).

Standard game theory \cite{osborne} defines a {\it game tree} as a graph in which nodes represent 
game states, with each branch corresponding to a move. The {\it complete} game tree for a problem is the game tree starting at the initial state and containing every possible move. Terminal nodes represent the possible states that may end the game; either a goal state or deadlock. The {\it branching factor} is the number of children at each node; an exhaustive search of the tree will follow every branch at every node and the total number of vertices will increase exponentially to the depth in the tree.

\begin{figure}[]
   \begin{center}
   \includegraphics[scale=0.35]{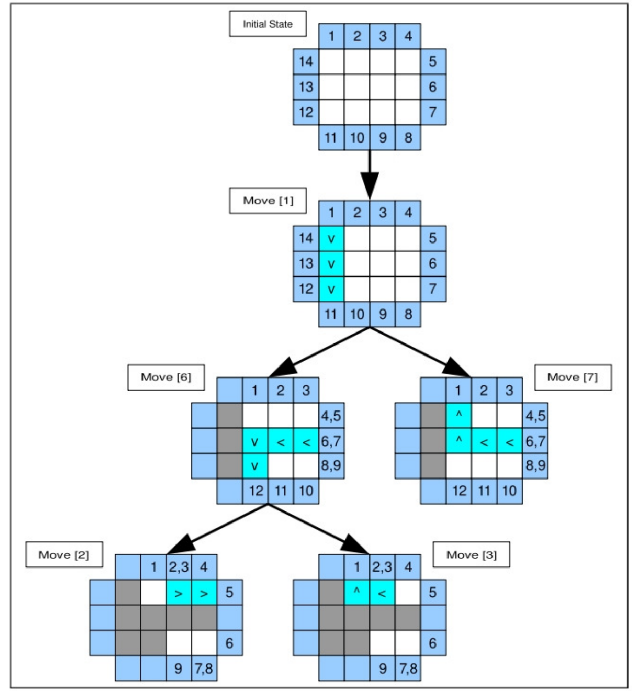}
       \caption{Partial first three ply for a simple ZPG board.}
       \label{fig:branch}
   \end{center}
 ~\\ 
 \end{figure}

A game tree is an example of state-space search whereby successive configurations
or states of an instance are considered, with the goal of finding a state with a desired
property. Figure  ~\ref{fig:branch} illustrates how this representation can be applied to the Zen Puzzle Garden. Obtaining a complete game tree for a problem can be very computationally expensive \cite{Hong}, and has resulted in search algorithms such as A* playing an important role in recent research.

\subsection{Simulator}

Attempts to solve commercial games are often hampered by the reluctance of the authors to release commercially-sensitive source code.  Therefore, in order to test candidate scripts against game boards, we use the same fundamental approach as Kendall and Spoerer\cite{Kendall}, which is to write a simplified version of the game engine. This engine retains the essential characteristics of the game in terms of its rules, but omits the graphical user interface and other ``playability" features. A representation of the game state is modified by an external controller program, and moves and their results may therefore be assessed without access to the source of the main game itself. 
Care is taken to ensure that all restrictions listed in Section 1 are enforced. The same simulator is used by both the genetic algorithm and informed search.

To determine the quality of a (partial) solution to a given garden, the simulator includes an $AreaFitness()$ method to calculate and return a value describing how close it is to a {\it full}solution. This method compares the unraked (available) area of the garden before and after a given path has been explored. By dividing the number of unraked tiles after the path has traversed the garden by the {\it initial} number of unraked tiles, a quality metric may be obtained. As the value approaches zero the proportion of the garden that
has been raked rises, with a value of zero indicating the path has covered all available tiles.

\subsection{A* solution}

The A* algorithm is an extensively studied best-first search method. It is best-first in
that it takes an {\it informed} approach when deciding which node is most likely to provide the least-cost distance to a goal state; the order in which nodes are visited is determined by a distance-plus-cost heuristic, given by $f(x) = g(x) + h(x)$, where:

\begin{itemize}
     \item $g(x)$: The shortest distance from the root node to the node being evaluated;
      \item $h(x)$: The estimated distance from the node being evaluated to a goal~state.
\end{itemize}

The heuristic function should be optimistic in that it will never overestimate the
cost of a path from the root node to a goal state; as the algorithm will never overlook
the possibility of a lower-cost path it is therefore admissible. Hart {\it et al.} \cite{hart} first
discussed this algorithm in 1968, then calling it just Algorithm A. It has since been
shown by Dechter and Pearl \cite{dechter} to be optimal in that it considers fewer nodes than any other admissible search algorithm with the same heuristic.

\begin{figure}[]
   \begin{center}
  \includegraphics[scale=0.2]{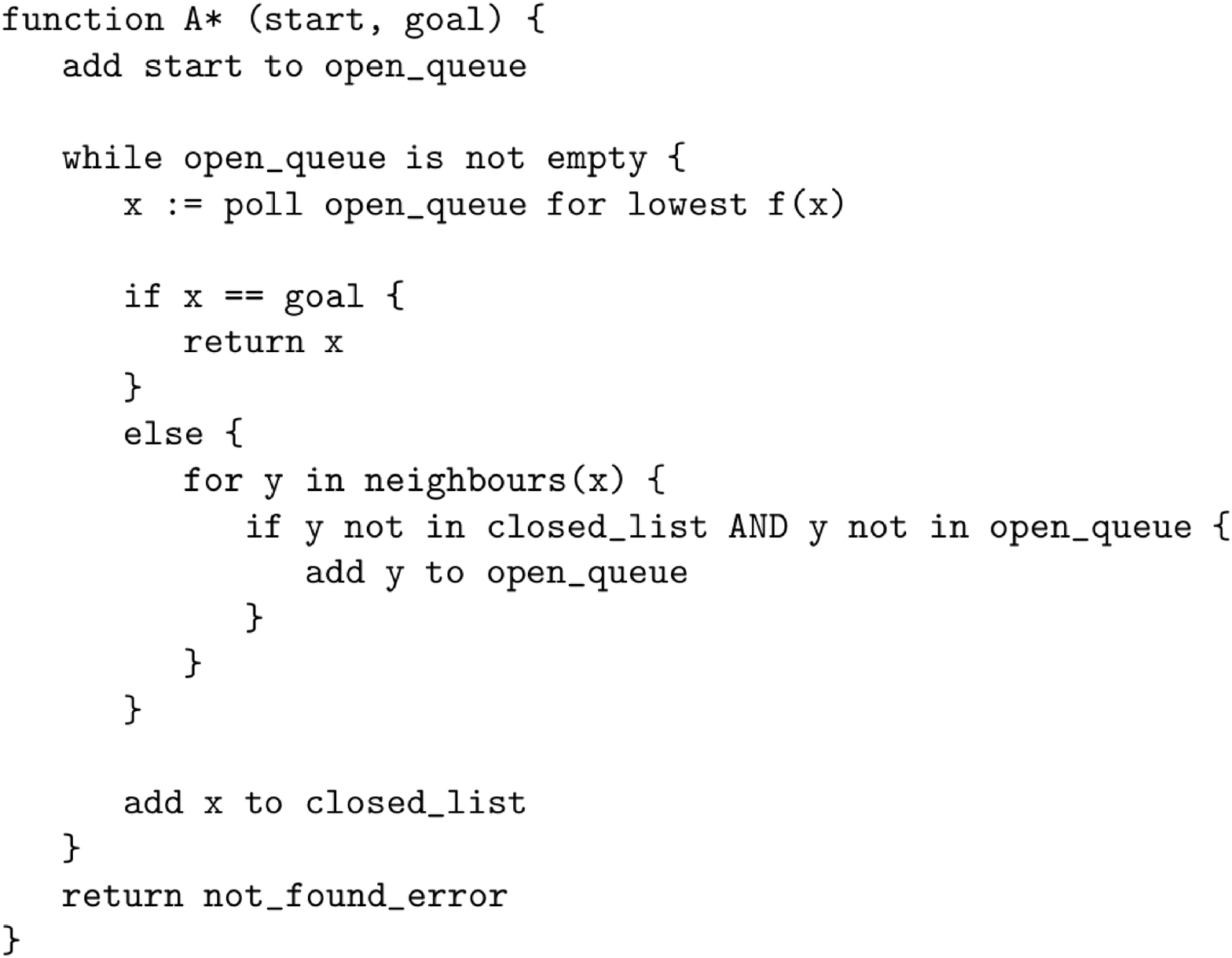}
       \caption{Pseudo-code for A* algorithm.}
       \label{fig:astar}
   \end{center}
 ~\\ 
 \end{figure}

The A* algorithm displays characteristics of the breadth-first search; it is complete
in that it will always return a solution if one exists, and it will also visit all equal-cost vertices at a given depth before continuing further along a path at greater cost. Many
enhancements have been made to the basic algorithm that allow it to be tailored
to the problem domain, such as iterative deepening and pattern databases. These
enhancements have provided some impressive reductions in the search effort required
to solve challenging problems such as Rubik's Cube \cite{korf}. The algorithm is implemented using a priority queue, the ordering determined by the heuristic function, with a lower value indicating a better candidate. The pseudo-code in Figure ~\ref{fig:astar} illustrates the basic operation of the algorithm.

The heuristic function comprises the distance
from the root node, $g(x)$, given by the number of moves completed, and the estimate
$h(x)$ is provided by the $AreaFitness()$ method described earlier.

The algorithm is implemented with a conditional loop which polls the fittest
state from the open queue and proceeds to create leaf nodes for each move available
from that state. As the nodes are created their quality is analysed, and they
are inserted into the queue accordingly. Once the first complete solution has been found
the queue is pruned of all nodes requiring a greater number of moves, and only
nodes with an move count less than or equal to the best solution are inserted into
the queue.

This process is repeated until all nodes in the queue have been examined, at which
point the solutions found are analysed and the set of unique solutions returned. Pruning reduces the completeness of the solution set generated by the algorithm for solutions requiring more than the optimal number of moves, but this
behaviour is acceptable, as only the optimal solutions are of interest in this investigation.

\subsection{Genetic algorithm solution}

Genetic algorithms \cite{goldberg89} have been studied extensively with respect to combinatorial optimisation problems. With respect to the current problem, Hong {\it et al.} \cite{Hong} discuss the application of a genetic algorithms to game search trees, specifically that of the Latin square problem devised by Leonhard Euler. Mantere and Koljonen \cite{mantere} perform a similar analysis of the efficacy of genetic algorithms for devising and solving Sudoku problems.

When choosing the representation for our GA, we first consider the basic operation of the game. A successful move consists of the selection of a point on the garden perimeter and a transition onto the garden surface, followed by zero or more in-move choices (the number of choices to be made is only non-zero if a raked square or object is encountered), leading to the monk finishing back on the perimeter. If a choice of direction is required at a raked square, then {\it at most} two options are available, as the monk can neither move backwards nor onto the raked square in front of him. If an ornament is encountered, then the monk must choose a number of times ($\geq 0$) to push it in the current direction. 

For a rectangular board of dimension $x \times y$, there are $C=2x+2y$ unique starting points on the perimeter, ordered by starting at 1 at the north face of the upper-left square and moving ``clockwise" to  the west face of the same square. $C$ is therefore a measure of the ``circumference" of the board. We assume that no move will contain more than 20 direction choices or opportunities to push. We define $m$ as the maximum number of moves allowed. A candidate solution is naturally made of a sequence of clauses 
\\ \\
\noindent
$(c_1, p_{1,1}, d_{1,1}, d_{1,1}, \dots, p_{1,8}, d_{1,8}), \dots \\ (c_i, p_{i,1}, d_{i,1}, \dots ,p_{i,8}, d_{i,8}), \dots \\ (c_m, p_{m,1}, d_{m,1}, \dots p_{m,8}, d_{m,8})$, 
\\ \\
\noindent
where each clause encodes a move,  $1 \leq c \leq C$, $1 \leq p_{i,j} < max(x,y)-1$ and $d_{i,j}  \in \{1,2\}$. 
Values for $c_i$ encode a starting point on the board perimeter. Values for $p_{i,j}$ encode a distance to push an object (if encountered). We use an indirect encoding scheme for values of $d_{i,j}$ to encode direction choices, using the notion of {\it available moves}. When a move is required, the list of available moves is constructed and one of two chosen (remembering that only two moves will be possible), according to the value of $d_i$. 
This genome sequence therefore encodes a ``script" of moves, which is then fed into the simulator for evaluation.

\subsubsection{Fitness Function}

The fitness function is given below. It accepts a sequence of moves, and returns the overall fitness of the sequence based on (a) its length, and (b) the quality of the solution it encodes. Fitness values are in the range $0 \dots 500$. $GENE\_LENGTH$ is equivalent to $m$, above, and is set to a default value of 20. $Moves$ represents the number of valid moves executed. The first component of the fitness function therefore rewards short sequences.

\begin{verbatim}
fitness()
{
  fitness=0 ;
  fitness += (GENE_LENGTH-Moves)*
                  (200/GENE_LENGTH)
  
  if (AreaFitness>0)
    fitness += (1-AreaFitness)*200
  else if (garden is full solution)
    fitness += 300
  else
    fitness = 0
}
\end{verbatim}

The second component of the fitness value is calculated by the $AreaFitness()$ function; a complete solution gains an absolute value of 300 for this component; if deadlock is reached (i.e., the monk is unable to move) a value of zero is awarded for this component.

\section{Results}
\label{results}

We tested both methods on 24 different ZPG game boards, including one engineered to have no solution. We selected nine ``retail" boards (supplied with the game, labelled R1-R9) and constructed an additional 14 ``test" boards (T1-T14). We could not simply select all retail boards, because we were restricted to using smaller boards due to the resource limitations we imposed (an informed search would be terminated if it either consumes more that 100Gb of disc space or runs for longer than 72 hours). Boards were selected/constructed in order to be ``tractable" in this sense, but we also ensured that the full range of board features was represented in our test suite\footnote{The board files, along with the code for the simulator, are available on request from the corresponding author.}

We initially ran both methods on 30 different ZPG game boards. Of these, 15 were ``retail" boards (i.e., supplied with the game), and 15 were hand-designed by us. Of the latter, one ``illegal" board was specifically engineered to have no solution, in order to demonstrate the exhaustive nature of the A* algorithm. The other 14 non-retail boards were designed to represent the range of obstacles contained in all 64 retail boards. Of the retail boards, eight were selected specifically because we expected them to exceed (due to their size) the A* resource limits that we impose; a search is terminated if it either consumes more than 100GB of disc cache space, or if it runs for longer than 72 hours. The 14 hand-designed boards were designed to be ``tractable" in this respect.

The A* algorithm was implemented in Java, using the JGAP \cite{meffert} Java package; it used a constant population size, rank-based selection (with 95\% of the population considered for the next generation),  and simple one-point crossover. The parameter values used were as follows: 

\begin{itemize}
\item Population size: 1000
\item Generations: 100
\item Chromosome length: 20
\item Mutation rate: 0.07
\end{itemize}

\begin{figure*}
   \begin{center}
   \includegraphics[scale=0.4]{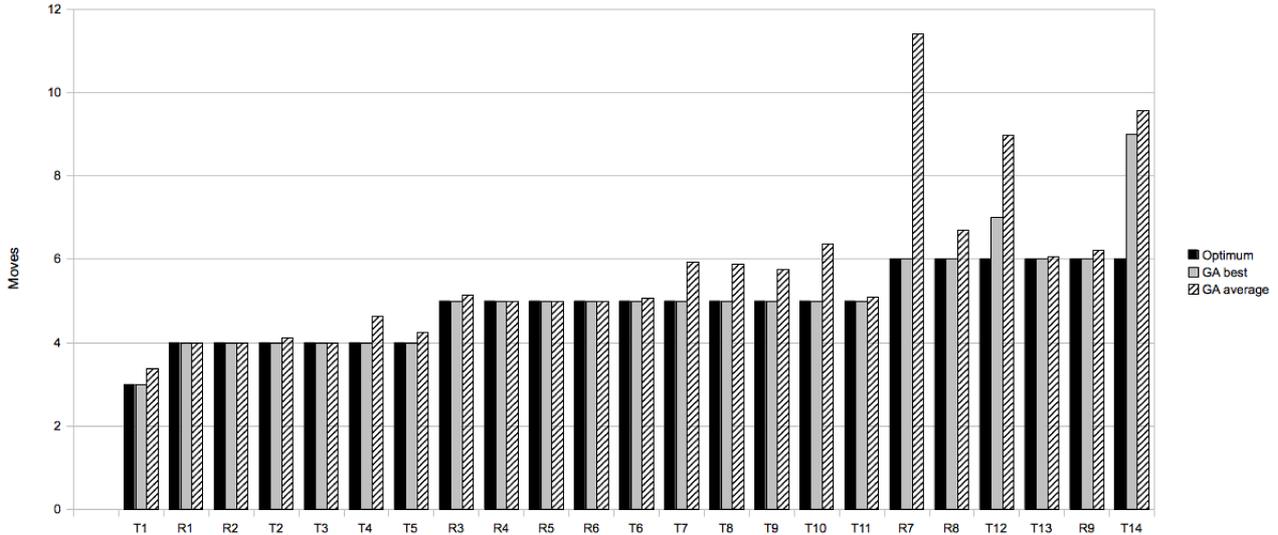}
       \caption{Graphical depiction of quality results.}
       \label{fig:results}
   \end{center}
 ~\\ 
 \end{figure*}

The GA was run 50 times on each board. Neither A* nor the genetic algorithm found valid solutions for the ``illegal" board, for which no solutions exist. The following results therefore describe comparisons over 23 ZPG boards. In Table 1, we first show the results in terms of solution quality. For each board, we give the the optimum number of moves for completion (as found by A*), the length of the {\it best} solution found by the GA, the {\it average} solution length for the GA, and the average GA ``quality overhead" in terms of excess moves. The table is ordered by the number of moves required by the optimal solution. We observe that the GA fails to find the optimal solution in only two cases (T12 and T14), and solves to optimality all of the retail boards tested (see graphical depiction of quality results in Figure ~\ref{fig:results}). Over all boards, on average the GA finds solutions that require roughly 15\% more moves than the optimal solution.

\begin{table}[]
\begin{centering}
\begin{tabular}{| c | c | c | c | c |}
\hline
{\bf Instance}	& {\bf Optimum}	& {\bf GA best}	& {\bf GA av.} 		& {\bf GA excess \%} \\
\hline
T1			& 3				& 3			& 3.38 			& 12.67 \\ \hline
R1			& 4				& 4			& 4 				& 0 \\ \hline
R2			& 4				& 4			& 4 				& 0 \\ \hline
T2			& 4				& 4			& 4.12 			& 3.0 \\ \hline
T3			& 4				& 4			& 4 				& 0 \\ \hline
T4			& 4				& 4			& 4.62 			& 15.5 \\ \hline
T5			& 4				& 4			& 4.24 			& 6.0 \\ \hline
R3			& 5				& 5			& 5.14 			& 2.8 \\ \hline
R4			& 5				& 5			& 5 				& 0 \\ \hline
R5			& 5				& 5			& 5 				& 0 \\ \hline
R6			& 5				& 5			& 5 				& 0 \\ \hline
T6			& 5				& 5			& 5.06 			& 1.2 \\ \hline
T7			& 5				& 5			& 5.94 			& 18.8 \\ \hline
T8			& 5				& 5			& 5.88 			& 17.6 \\ \hline
T9			& 5				& 5			& 5.76 			& 15.2 \\ \hline
T10			& 5				& 5			& 6.36 			& 27.2 \\ \hline
T11			& 5				& 5			& 5.08 			& 1.6 \\ \hline
R7			& 6				& 6			& 11.41			& 90.17 \\ \hline
R8			& 6				& 6			& 6.7 			& 11.67 \\ \hline
T12			& 6				& 7			& 8.98 			& 49.67 \\ \hline
T13			& 6				& 6			& 6.06 			& 1.0 \\ \hline
R9			& 6				& 6			& 6.22			& 3.67 \\ \hline
T14			& 6				& 9			& 9.57 			& 59.5 \\ \hline
			&				&			& {\bf Average}		& {\bf 14.66} \\ \hline
\end{tabular}
\caption{Move count comparison for A* and GA.}
\end{centering}
\label{tab:quality}
\end{table}

We now consider the computational effort required by each algorithm. We measure this in terms of {\it board evaluations required}, as this is the only truly objective metric. In addition, in order to ensure a fair comparison we only consider boards where the GA finds the {\it optimal solution}, so we discount boards T12 and T14. The results are given in Table ~\ref{tab:runtime}. The second column gives the number of boards evaluated by A* to find the optimal solution. The third column gives the average number of fitness evaluations required by the GA to find the optimal solution, and the fourth shows this number as a proportion of the number of A* evalutions (i.e., anything less than 100 shows an improvement over A*).

\begin{table}
\begin{centering}
\begin{tabular}{|c|c|c|c|c|}
\hline
{\bf Instance}	& {\bf A* evals.}&  {\bf Av. GA evals.}	& {\bf \% of A*} \\
\hline

T1	& 903,634		& 27000		& 2.99 	\\ \hline
R1	& 123,084		& 1000		& 0.81	\\ \hline
R2	& 152,265		& 1000		& 0.66	\\ \hline
T2	& 77,272		& 30000		& 38.82	\\ \hline
T3	& 109,284		& 1000		& 0.92	\\ \hline
T4	& 2,962,349	& 48000		& 1.62	\\ \hline
T5	& 956,861		& 35000		& 3.66	\\ \hline
R3	& 30,759,145	& 25000		& 0.08	\\ \hline
R4	& 2,983,478	& 1000		& 0.03	\\ \hline
R5	& 2,983,134	& 2000		& 0.07	\\ \hline
R6	& 912,676		& 1000		& 0.11	\\ \hline
T6	& 1,137,751	& 13000		& 1.14  	\\ \hline
T7	& 158,756,106	& 50000		& 0.03  	\\ \hline
T8	& 15,092,790	& 45000		& 0.3	\\ \hline
T9	& 33,414,980	& 53000		& 0.16	\\ \hline
T10	& 128,639,420	& 47000		& 0.04	\\ \hline
T11	& 9,896,968	& 32000		& 0.32	\\ \hline
R7	& 5,888,672	& 31000		& 0.53	\\ \hline
R8	& 15,728,328	& 48000		& 0.31	\\ \hline
T13	& 22,742,096	& 12000		& 0.05	\\ \hline
R9	& 5,888,672	& 31000		& 0.53	\\ \hline
      	&			& {\bf Average}	& {\bf 2.53}	\\ \hline
\end{tabular}
\caption{Number of evaluations for A* and GA.}
\label{tab:runtime}
\end{centering}
\end{table}

We observe that, with the exception of a single problem instance (T2), compared to informed search the GA requires significantly fewer evaluations to find the optimum. Over all boards, the GA requires, on average, under 3\% of the evaluations needed by A*. If we discount board T2, this falls to 0.65\%. 

\begin{figure}
   \begin{center}
   \includegraphics[scale=0.4]{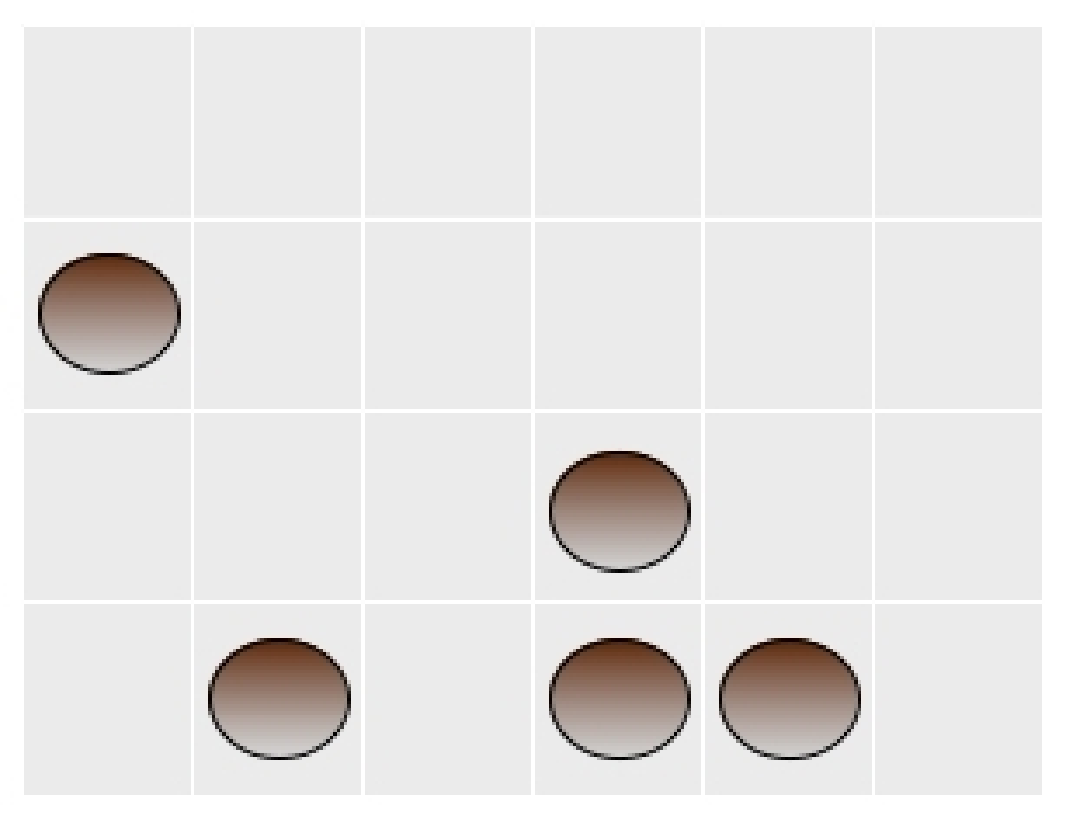}
       \caption{The ``problematic" board T2.}
       \label{fig:t2}
   \end{center}
 ~\\ 
 \end{figure}

Board T2 contains only rocks, and it is not immediately apparent from inspection why the GA struggles so much with this particular instance. The board is shown in Figure ~\ref{fig:t2}; we note its small size, and relatively dense population of rocks (21\% of the board is occupied at the outset). We believe that this density places significant constraints on possible solutions, which is why this board has by far the smallest search space. It may be that, in cases such as these, it is more efficient to solve the instance using informed search. 

\section{Conclusions}

In this paper we have described a novel genetic algorithm solution to a block-based puzzle game. The game poses significant challenges in terms of the size of its search space, but our solution is competitive with informed search in terms of solution quality, and significantly out-performs it in terms of its computational resource requirements.

We presented the ZPG in order to both demonstrate the efficacy of a genetic algorithm solution, and to encourage further study of its properties. Inspired by \cite{kendall2}, we wish to investigate questions such as ``Is it possible to automatically generate hard and easy instances of the problem?", as well as considering the notion of an {\it aesthetically pleasing} solution.
In addition to providing a useful testbed for new solution methods, the problem domain has real significance if we consider the problem of mobile robotics, where a self-avoiding path must be chosen whilst also considering possible obstacles and moveable objects. Future work will focus on formally establishing the difficulty of the ZPG game, as well as further investigations into its solution.

\section*{Acknowledgments}

The authors thank Joseph White (author of the ZPG) for invaluable assistance with
the game, David Corne for useful advice on representation
schemes, and David Goldberg \cite{zen} and Robert M. Pirsig \cite{zen2} for titular inspiration.

\end{document}